\documentclass{article}

\usepackage[preprint,nonatbib]{nips_2018}

\usepackage[utf8]{inputenc} 
\usepackage[T1]{fontenc}    
\usepackage{hyperref}       
\usepackage{url}            
\usepackage{booktabs}       
\usepackage{amsfonts}       
\usepackage{amsmath}
\usepackage{nicefrac}       
\usepackage{microtype}      
\usepackage{graphicx}
\usepackage{algpseudocode}
\usepackage{algorithmicx,algorithm}
\usepackage{bm}
\usepackage{subfig}

\title{Cross-City Transfer Learning for Deep Spatio-Temporal Prediction}

\author{
	Leye Wang\thanks{Equal contribution} \\
	HKUST \\
	\texttt{wly@cse.ust.hk}\\
	\And
	Xu Geng$^*$\\
	HKUST\\
	\texttt{xgeng@connect.ust.hk}\\
	\And
	Xiaojuan Ma\\
	HKUST\\
	\texttt{mxj@cse.ust.hk}\\
	\And
	Feng Liu\\
	SAIC Motor\\
	\texttt{liufeng@saicmotor.com}\\
	\And
	Qiang Yang\\
	HKUST\\
	\texttt{qyang@cse.ust.hk}
}

\begin{document}

\maketitle

\begin{abstract}
  Spatio-temporal prediction is a key type of tasks in urban computing, e.g., traffic flow and air quality. Adequate data is usually a prerequisite, especially when deep learning is adopted. However, the development levels of different cities are unbalanced, and still many cities suffer from data scarcity. To address the problem, we propose a novel cross-city transfer learning method for deep spatio-temporal prediction tasks, called \textit{RegionTrans}. \textit{RegionTrans} aims to effectively transfer knowledge from a data-rich source city to a data-scarce target city. More specifically, we first learn an inter-city region matching function to match each target city region to a similar source city region. A neural network is designed to effectively extract region-level representation for spatio-temporal prediction. Finally, an optimization algorithm is proposed to transfer learned features from the source city to the target city with the region matching function. Using citywide crowd flow prediction as a demonstration experiment, we verify the effectiveness of \textit{RegionTrans}. Results show that \textit{RegionTrans} can outperform the state-of-the-art fine-tuning deep spatio-temporal prediction models by reducing up to 10.7\% prediction error.
  
\end{abstract}

\section{Introduction}

Spatio-temporal prediction covers a broad scope of applications in urban computing \cite{zheng2014urban}, such as traffic and air quality prediction. Recently, with the development of big data techniques, deep learning becomes popular in spatio-temporal prediction, e.g. crowd flow, taxi demand, precipitation predictions, and achieves state-of-the-art performance \cite{ke2017short,xingjian2015convolutional,zhang2017deep,zhang2016dnn}. However, the city development levels are quite unbalanced, so that many cities cannot benefit from such achievements due to data scarcity. Hence, how to help data-scarce cities also obtain benefits from the recent technique breakthroughs like deep leaning, becomes an important research issue, while it is still under-investigated up to date.

To tackle this problem, in this paper, we propose a new cross-city transfer learning method for deep spatio-temporal prediction tasks, called \textit{RegionTrans}. The objective of \textit{RegionTrans} is to predict a certain type of service data (e.g., crowd flow) in a data-scarce city (\textit{target city}) by transferring knowledge learned from a data-rich city (\textit{source city}). The principal idea of \textit{RegionTrans} is to find \textit{inter-city region pairs} that share similar patterns and then use such region pairs as proxies to efficiently transfer knowledge from the source city to the target. 

%
%


In literature, existing deep learning approaches are often designed to predict citywide phenomenon as a whole \cite{zhang2017deep,zhang2016dnn}, and thus it is hard to enable region-level knowledge transfer. To this end, rather than adopting the existing deep neural networks for citywide spatio-temporal prediction (e.g.  ST-ResNet \cite{zhang2017deep}), we propose a novel deep transfer learning method. First, we design a region matching function to link each target city region to a similar source region based on the short period of service data or correlated auxiliary data if applicable. Then, in our proposed network structure, to catch the spatio-temporal patterns hidden in the service data, ConvLSTM layers \cite{xingjian2015convolutional} are firstly stacked. Afterward, to encode \textit{region representation}, we newly add a Conv2D layer with $1 \times 1$ filter, which is the key and fundamental component of our network to make region-level transfer feasible. Finally, the discrepancy between \textit{region representations} of the inter-city similar regions is minimized during the network parameter learning for the target city, so as to enable region-level cross-city knowledge transfer. With crowd flow prediction as a showcase~\cite{zhang2017deep,zhang2016dnn}, we verify the feasibility and effectiveness of \textit{RegionTrans}. Briefly, this paper has the following contributions.

(i) To the best of our knowledge, this is the first work to study how to facilitate deep spatio-temporal prediction in a data-scarce target city by transferring knowledge from a data-rich source city.

(ii) We propose a novel  deep transfer learning method \textit{RegionTrans} for spatio-temporal prediction tasks by region-level cross-city transfer. \textit{RegionTrans} first computes inter-city region similarities, and then stacks ConvLSTM and Conv2D ($1\times1$ filter) layers to extract region-level representations reflecting spatio-temporal patterns. Finally, the discrepancy of the representations of inter-city similar regions is minimized so as to facilitate region-level cross-city knowledge transfer.


(iii) With crowd flow prediction as a showcase, our experiment shows that \textit{RegionTrans} can reduce up to 10.7\% prediction error compared to fine-tuned state-of-the-art spatio-temporal prediction methods.
\section{Related Work}

\textit{Spatio-Temporal Prediction} is a fundamental problem in urban computing \cite{zheng2014urban}.  
Recently, deep learning is adopted in spatio-temporal prediction tasks and becomes the state-of-the-art solution when there exists a rich history of data. Various deep models have been used, e.g., CNN \cite{zhang2016dnn}, ResNet~\cite{zhang2017deep}, and ConvLSTM \cite{ke2017short,xingjian2015convolutional,yao2018deep}. Compared to these works, the  difference of our work lies in both objective and method. We aim to apply deep learning to a target city with a short period of service data, and thus propose \textit{RegionTrans} to effectively transfer knowledge from a data-rich source city to the target city.

\textit{Transfer Learning} addresses the machine learning problem when labeled training data is scarce \cite{pan2010survey}. In urban computing, data scarcity problem often exists when the targeted service or infrastructure is new. There are generally two strategies to deal with urban data scarcity. The first is using auxiliary data of the target city to help build the targeted application. Examples include using temperature to infer humidity and vice versa \cite{wang2017space}, and leveraging the taxi GPS traces to detect ridesharing cars \cite{wang2017ridesourcing}. The second is to find a source city with adequate data to transfer knowledge.  Guo et al. design a cross-city transfer learning framework with collaborative filtering and AutoEncoder to conduct chain store site recommendation \cite{Guo2017City}. As our problem is prediction rather than recommendation, the method in \cite{Guo2017City} cannot be applied. Another relevant work is \cite{wei2016transfer}, which proposes a cross-city transfer learning algorithm FLORAL to predict air quality category. There are two difficulties to apply FLORAL to our task: (1) many spatio-temporal prediction tasks are regression but FLORAL is designed for classification; (2) FLORAL is not designed for deep learning. As far as we know, \textit{RegionTrans} is the first cross-city transfer learning framework for deep spatio-temporal prediction.

\section{Problem Formulation}

\textbf{Definition 1. Region}~\cite{zhang2016dnn}. A city $\mathcal D$ is partitioned into $W_\mathcal D \times H_\mathcal D$ equal-size grids (e.g., $1 km \times 1 km$). Each grid is called a \textit{region}, denoted as $r$. We use $r_{[i,j]}$ to represent a city region whose coordinate is $[i,j]$. The whole set of regions in a city $\mathcal D$ is denoted as $\mathbb C_{\mathcal D}$.

\textbf{Definition 2. Urban Image Time Series}. We denote the \textit{set of data time-stamps} of a city $\mathcal D$ as:
\begin{equation}
	\mathbb T_{\mathcal{D}} = [t_c-T_{\mathcal{D}}+1, t_c]
\end{equation}
where $T_{\mathcal{D}}$ is the number of time-stamps and $t_c$ is the current/last time-stamp. For brevity, we consider equal-length time-stamp (e.g., one-hour) as in the previous research \cite{zhang2017deep,zhang2016dnn}.
For a specific time-stamp $t$, we have an \textit{urban image} $\mathcal I_{t, \mathcal D}$ with $W_\mathcal D \times H_\mathcal D$ pixels where each pixel represents certain data  of a corresponding region (Def. 1),
\begin{equation}
	\mathcal I_{t,\mathcal D} = \{i_{r,t}|r \in \mathbb C_{\mathcal D}\} \in \mathbb R^{W_\mathcal D \times H_\mathcal D}
\end{equation}
Then, we define an \textit{urban image time series} $\mathbb I_{\mathcal D}$ as follows:
\begin{equation}
	\mathbb I_\mathcal D = \{\mathcal I_{t,\mathcal D}|t\in \mathbb T_{\mathcal D} \}  \in \mathbb R^{T_{\mathcal D} \times W_\mathcal D \times H_\mathcal D}
\end{equation}
In reality, a variety of urban data can be modeled as the above urban image time series, such as crowd flow, weather condition, air quality, etc. 

\textbf{Definition 3. Service Spatio-temporal Data}. Service data is the targeted type of data to predict.
We define the service spatio-temporal data as the urban image time series $\mathbb S^\mathcal D$ storing the service data :
\begin{equation}
\mathbb S_{\mathcal D} = \{ \mathcal S_{t,\mathcal D} | t \in \mathbb{T}_{\mathcal D} \} = \{s_{r,t}|r \in \mathbb C_\mathcal D, t \in \mathbb T_{\mathcal D} \} \in \mathbb R^{T_{\mathcal D} \times W_\mathcal D \times H_\mathcal D}
\end{equation}
where $s_{r,t}$ is the service data of region $r$ at time-stamp $t$.


In this paper, the target city $\mathcal D$ suffers from the service data scarcity, while the source city $\mathcal D'$ has rich service data, i.e., $|\mathbb T_{\mathcal D}| \ll |\mathbb T_{\mathcal D'}|$. 
With this in mind, we formulate the problem.

\textbf{Problem. Spatio-temporal Prediction by Cross-city Transfer}. 
Given the little service data in target city $\mathcal D$ and rich service data in source city $\mathcal D'$, we aim to learn a function $f$ to predict the citywide service data in the target city $\mathcal D$ at the next time-stamp $t_c+1$:
\begin{align}
	& \min_f \quad \textit{error}( \mathcal{\widetilde S}_{t_c+1, \mathcal D}, \mathcal{S}_{t_c+1, \mathcal D})\\
	 \text{where} \quad & \mathcal{\widetilde S}_{t_c+1, \mathcal D} = f(\mathbb{S}_{\mathcal D},\mathbb{S}_{\mathcal D'}), \quad |\mathbb T_{\mathcal D}| \ll |\mathbb T_{\mathcal D'}| 
\end{align}
The \textit{error} metric may be mean absolute error, root mean squared error, etc., according to the real application requirement.

\textbf{Example. Crowd Flow Prediction}. We use crowd flow prediction \cite{zhang2017deep,zhang2016dnn} as an example to illustrate the above problem concretely. The service data $\mathbb S_{\mathcal D}$ is thus crowd inflow or outflow.  The  source city crowd flow records may last for several years ($\mathbb T_{\mathcal D'}$), but the target city may have only a few days ($\mathbb T_{\mathcal D}$) as the service is just started. 
It is worth noting that external context factors, such as weather and workday/weekend, are also important in crowd flow prediction \cite{zhang2017deep}. Later we will show that our proposed method is easy to add the external features extracted from context factors.


\section{RegionTrans}

To solve the above problem, we propose a deep transfer learning method \textit{RegionTrans} including three stages. First, we learn an inter-city region matching function that links each target region to a similar source region. Second, a neural network structure is designed to extract region-level spatio-temporal patterns. Finally, an optimization process is proposed to facilitate region-level transfer between cities.

\subsection{Inter-city Similar-region Matching}
\label{sub:region_match}

The first step of \textit{RegionTrans} is to find a matching function $\mathcal M: \mathbb C_{\mathcal D} \to \mathbb C_{\mathcal D'}$ to map each region of the target city $\mathcal D$ to a certain region of the source city $\mathcal D'$. The objective is to find the source region having the similar spatio-temporal pattern with the target region. To this end, we propose two strategies to find $\mathcal M$.

\textbf{Matching with a short period of service data}. While the target city has only a little service data, this could still provide hints to build $\mathcal M$. We focus on the time span when both source and target cities have service data (i.e., $\mathbb T_{\mathcal D}$), then calculate the correlations (e.g., \textit{Pearson} coefficient) between each target region and source region with the corresponding service data. Finally, for each target region, we choose the source region with the largest correlation value. Formally,
\begin{align}
	&\mathcal M(r) = r^*, & \ r \in \mathbb C_{\mathcal D}, r^*\in \mathbb C_{\mathcal D'} \\
	&\rho_{r,r^*} \ge  \rho_{r,r'}, & \forall r' \in \mathbb C_{\mathcal D'} \\
	&\rho_{r,r^*} = corr(\{s_{r,t}\}, \{s_{r^*,t}\}), & \ r \in \mathbb C_{\mathcal D}, r^*\in \mathbb C_{\mathcal D'} , t \in \mathbb T_{\mathcal D}
	\label{eq:service_match}
\end{align}
\textbf{Matching with a long period of auxiliary data (if applicable)}. As there is little service data in the target city, the above service-data-based correlation similarity between a source region and a target region may not be very reliable. In reality, sometimes we can find another openly-accessible auxiliary data that correlates with the service data, which may help calculate the inter-city region similarity more robustly. For example, to predict crowd flow, public social media check-ins can be a useful proxy according to literature~\cite{yang2016participatory}. That is, instead of the short period of crowd flow data, we use the long period of openly available check-in data to build the correlation between two regions.
\begin{align}
	&\rho_{r,r^*} = corr(\{a_{r,t}\}, \{a_{r^*,t}\}), & \ r \in \mathbb C_{\mathcal D}, r^*\in \mathbb C_{\mathcal D'} , t \in \mathbb T_{\mathcal A}
\end{align}
where $a$ is the auxiliary data (e.g., check-in number) lasting for a long period $\mathbb T_{\mathcal A} (|\mathbb T_{\mathcal A}| \gg |\mathbb T_{\mathcal D}|)$.

\subsection{Deep Spatio-temporal Neural Network with Region Representations}
\label{sub:network_structure}
\begin{figure}[t]
	\centering
	\includegraphics[width=.88\linewidth]{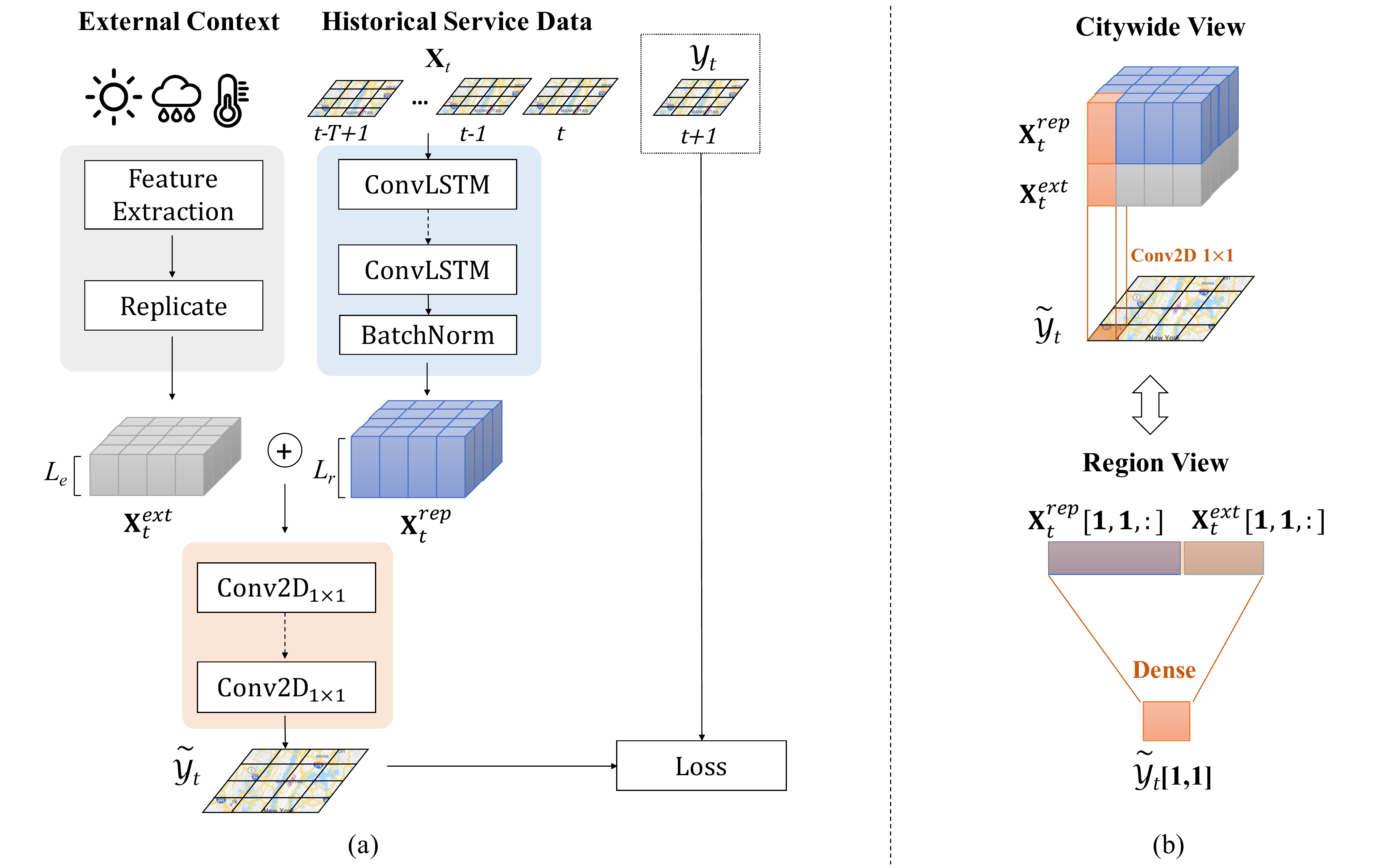}
	\caption{Proposed network structure.}
	\label{fig:network}
	\vspace{-1em}
\end{figure}

Existing deep spatio-temporal models often take the whole city data for end-to-end prediction (e.g., ST-ResNet \cite{zhang2017deep}), which cannot be used for region-level transfer. Therefore, we design a new network for spatio-temporal prediction with region representations, as shown in Fig.~\ref{fig:network}~(a).

\textbf{Input and output}. First we illustrate the input and output of the proposed network\footnote{For clarity, we omit the subscript $\mathcal D$ in notations as all the notations mentioned in this section is in city $\mathcal D$.}:
\begin{align}
	&k \in \mathbb{N}^+ & \text{as the length of the input time series} \\
	&\mathbf{X}_t = \{\mathcal{S}_{t'} | t' \in [t-k+1, t]\} \in \mathbb{R}^{k\times W\times H}& \text{as input for prediction}\\
	&\mathcal{Y}_{t} = \mathcal{S}_{t+1} \in \mathbb{R}^{W\times H} & \text{as ground-truth result at time } t+1\\
	&f_{\theta} : \mathbb{R}^{k\times W\times H} \rightarrow \mathbb{R}^{W\times H}& \text{as neural network with parameter }\theta \\
	& \widetilde{\mathcal{Y}}_{t} = f_{\theta}(\mathbf{X}_t) \in \mathbb{R}^{W\times H} & \text{as prediction result at time } t+1
\end{align}
 Our network objective is to minimize the squared error between  predicted $\widetilde{\mathcal{Y}}_{t}$ and real $\mathcal{Y}_{t}$:
\begin{equation}
	 \min_\theta \sum_{t \in \mathbb T} || \widetilde{\mathcal{Y}}_{t} - \mathcal{Y}_{t} ||^2_F
\end{equation}
\textbf{Network structure.} ConvLSTM layers are used as the basic components for our neural network to learn spatio-temporal patterns \cite{xingjian2015convolutional}. In the first part of the network, we use a set of stacked ConvLSTM layers to construct region-level hidden representation $\mathbf{X}_t^{rep} \in \mathbb{R}^{W\times H\times L_{r}}$ as defined in Eq.~\ref{convlstm} (we will elaborate why this can be seen as region-level representation soon). 
After getting $\mathbf{X}_t^{rep}$, we incorporate the external context factors into the network structure. External context factors are defined as $\mathbf X_t^{ext} \in \mathbb{R}^{W\times H\times L_{e}}$, which is a feature vector of length $L_{e}$ on each region (e.g., weather, temperature, weekday/holiday one-hot encoding \cite{zhang2017deep}). By concatenating $\mathbf{X}_t^{rep}$ and $\mathbf{X}_t^{ext}$ to form a representation $\in \mathbb R^{W\times H \times (L_r+L_e)}$, we employ several convolution 2D layers with $1\times 1$ filters (Conv2D$_{1\times 1}$ \cite{lin2014network}) to predict the next-time-stamp service data $\widetilde{\mathcal{Y}}_{t} \in \mathbb R^{W \times H}$. Formally, the spatio-temporal neural network can be formulated as follows:
\begin{align}
	&  f_{\theta_{1}}: \mathbb{R}^{k\times W\times H} \rightarrow \mathbb{R}^{W\times H\times L_{r}} & \text{as ConvLSTM layers} \label{convlstm} \\
	&  f_{m}: (\mathbb{R}^{W\times H\times L_{r}},\mathbb{R}^{W\times H\times L_{e}}) \rightarrow \mathbb{R}^{W\times H\times (L_{r}+L_{e})} & \text{as merge layer} \\
	&  f_{\theta_{2}}: \mathbb{R}^{W\times H\times (L_{r}+L_{e})} \rightarrow \mathbb{R}^{W\times H}& \text{as Conv2D$_{1\times 1}$ layers}\\
	&  \mathbf X_t^{rep} = f_{\theta_1}(\mathbf X_t) & \text{as region representation} \\
	& \widetilde{\mathcal Y}_{t} = f_{\theta_2}(f_m(\mathbf X^{rep}_t, \mathbf X_t^{ext})) = f_{\theta_2}(f_m(f_{\theta_1}(\mathbf X_t), \mathbf X_t^{ext})) &\text{as prediction output}
\end{align}


\textbf{Region representation.} As visualized by Fig.~\ref{fig:network}~(b), Conv2D$_{1\times 1}$ will produce spatio-invariant results, which means hidden vector $\mathbf X^{rep}_t[w,h,:]$ and prediction $\widetilde{\mathcal Y}_{t}[w,h]$ represent the spatio-temporal representation and prediction result of region $r_{[w,h]}$, respectively.
Compared with existing end-to-end citywide deep spatio-temporal prediction models \cite{zhang2017deep,zhang2016dnn} without such region-level hidden representations, our network design has the following advantages for transfer learning:

(i) \textit{Fine-grained region-level transfer}. 
With existing methods which consider the data of a city as a whole for prediction, we can only transfer the knowledge from the whole source city to the target (e.g., through fine-tuning). If two cities are not similar in general, the transfer performance may be poor. As our network incorporates region representation, we can make fine-grained knowledge transfer based on region similarity (the detailed algorithm in the next sub-section). As long as we can find similar region pairs between cities, the effective transfer may be conducted.

(ii) \textit{Transfer between cities with different sizes}. Since our neural network structure can be seen from region view (Fig.~\ref{fig:network}~(b)), even if two cities have different sizes (i.e. $W, H$), it is possible to train a model on a source city and then transfer the learned network parameters to the target city at the region level. However, with end-to-end network structures \cite{zhang2017deep,zhang2016dnn}, if we want to transfer a learned model from the source city to the target by fine-tuning, the two cities must be the same size.

\subsection{Region-based Cross-city Network Parameter Optimization}

With the proposed network structure, we train a deep model in the source city $\mathcal D'$ with its rich spatio-temporal service data. We denote $\theta_{\mathcal D'}$ as the network parameters learned from the source city. Then, with $\theta_{\mathcal D'}$ as the pre-trained network parameters, we propose a region-based cross-city optimization algorithm to refine the network parameters on the target city $\mathcal D$, considering a short period $\mathbb T_{\mathcal D}$ of the service data in the target city $\mathcal D$ and the inter-city region matching function $\mathcal M$. 

When refining the network parameter for the target city $\mathcal D$, the first objective is to minimize prediction error on the target city:
\begin{equation}
	\min_{\theta_{\mathcal D}}  \sum_{t \in \mathbb T_{\mathcal D}} || \widetilde{\mathcal{Y}}_t - \mathcal{Y}_t ||^2_F
\end{equation}
Given the matching function $\mathcal M$, the second objective of our optimization is to minimize representation divergence between matched region pairs. More specifically, for each time-stamp $t \in \mathbb T_{\mathcal D}$, we try to minimize the squared error between the network hidden representations of the target region and its matched source region.
Formally, the second objective is as follows:
\begin{align}
\label{eq:obj2}
\min_{\theta_{\mathcal D}}\quad \sum_{r \in \mathbb C_{\mathcal D}} \sum_{t \in \mathbb T_{\mathcal D}} \rho_{r,r^*} \cdot ||\mathbf{x}^{rep}_{r,t}-\mathbf{x}^{rep}_{r^*,t}||^2, \text{ where } r^* = \mathcal M(r)
\end{align}
where $\mathbf{x}^{rep}_{r,t}$ is the hidden representation of the target region $r$ when the last input time-stamp is $t$; $\mathbf{x}^{rep}_{r^*,t}$ is the representation of the matched source region $r^*$; $\rho_{r,r^*}$ is the correlation value calculated between region pairs (Sec.~\ref{sub:region_match}), so that more similar pair will be assigned with larger weights in the optimization. 
Then, combining the two objectives leads to the following optimization process:
\begin{equation}
\label{eq:obj}
\begin{aligned}
	 \min_{\theta_{\mathcal D}} \quad & (1-w) \sum_{t \in \mathbb T_{\mathcal D}} || \widetilde{\mathcal{Y}}_t - \mathcal{Y}_t ||^2_F + w \sum_{r \in \mathbb C_{\mathcal D}} \sum_{t \in \mathbb T_{\mathcal D}} \rho_{r,r^*} \cdot ||\mathbf{x}^{rep}_{r,t}-\mathbf{x}^{rep}_{r^*,t}||^2
\end{aligned}
\end{equation}
where $w$ is the weight to trade off between minimizing the representation discrepancy or minimizing the prediction error. Then, we can use state-of-the-art network parameter learning algorithms, such as \textit{SGD} and \textit{ADAM}, to obtain the network parameter $\theta_{\mathcal D}$ for the target city $\mathcal D$ according to Eq.~\ref{eq:obj} (the network parameter $\theta_{\mathcal D'}$ learned in the source city $\mathcal D'$ is used as the initialization values). The detailed pseudo-code of the optimization process is summarized in Alg.~\ref{alg:regiontrans}.

\begin{algorithm}[t]
	\scriptsize 
	\caption{Region-based cross-city network parameter optimization} 
	\label{alg:regiontrans}
	{\bf Input:} \\
	\hspace*{0.02in} $\theta_{\mathcal D'}$: Pre-trained network parameters on source city with a long period of service data \\
	\hspace*{0.02in} $TR_{\mathcal D}$: target city training data \\
	\hspace*{0.02in} $TR_{\mathcal D'}$: source city training data \\
	\hspace*{0.02in} $\mathcal M$: inter-city similar-region matching function\\
	{\bf Output:} \\
	\hspace*{0.02in} $\theta_{\mathcal D}$: network parameters for the target city\\
	
	\begin{algorithmic}[1]
		\State Initialize network parameters: $\theta \leftarrow \theta_{\mathcal D'}$
		\State epoch $ \leftarrow 0$
		\While{epoch $\le$ MAX\_EPOCH} 
		
		\For {$t \in \mathbb T_{\mathcal D}$}
		\State Get $\{\mathbf{X}_{t}, \mathcal{Y}_{t} \} \in TR_{\mathcal D} $
		\State Get corresponding $\{\mathbf{X}'_{t},  \mathcal{Y}'_{t} \} \in TR_{\mathcal D'} $
		\For{$r \in \mathbb{C}_{\mathcal D}$}
		\State$r^* \leftarrow \mathcal M(r)$ (note that $r^* \in \mathbb{C}_{\mathcal D'}$)
		\State$\mathbf{x}^{rep}_{r^*,t} \leftarrow$ region representation of network ($\theta$) with input  $\mathbf{X}'_{t}$ for source region $r^*$
		\State$\mathbf{x}^{rep}_{r,t} \leftarrow$ region representation of network ($\theta$) with input  $\mathbf{X}_{t}$ for target region $r$
		\EndFor
		\EndFor
		\State \vspace{-1.4em}
		\begin{flalign*}
		\quad \ \ \ \theta \leftarrow \arg \min_{\theta} \quad (1-w) \sum_{t \in \mathbb T^{\mathcal D}} || \widetilde{\mathcal{Y}}_t - \mathcal{Y}_t ||^2_F + w \sum_{r \in \mathbb C^{\mathcal D}} \sum_{t \in \mathbb T^{\mathcal D}} \rho_{r,r^*} \cdot ||\mathbf{x}^{rep}_{r,t}-\mathbf{x}^{rep}_{r^*,t}||^2 &&
		\end{flalign*}
		\State epoch ++
		\EndWhile
		\State $\theta_{\mathcal D} \leftarrow \theta$
		\State \Return $\theta_{\mathcal D}$
	\end{algorithmic}
\end{algorithm}

\subsection{Remark: Divide-Match-Transfer Principle}
Here, we elaborate the key principle behind \textit{RegionTrans}, which we term as \textit{divide-match-transfer}. That is, for the target city domain and the source city domain, instead of transferring knowledge from the source to the target as a whole, we first \textit{divide} both domains into a set of regions, or called \textit{sub-domains}. Then, we build a \textit{matching} between the target sub-domains and the source sub-domains. Finally, with the matched cross-sub-domain pairs, we conduct the knowledge \textit{transfer}. Previous theoretical transfer learning studies \cite{ben2007analysis,blitzer2008learning,DBLP:conf/colt/MansourMR09} have proved that the feature distribution difference between the source and target domains is a key factor impacting the transfer learning performance. Then, if we can build a reasonable cross-sub-domain matching (i.e., the feature distribution difference between the two matched sub-domains becomes smaller than the original two domains), it is probable that we can improve the transfer learning performance. As this is an intuitive understanding of the \textit{divide-match-transfer} principle, we will theoretically study its properties in the future work.

With \textit{divide-match-transfer} principle  in mind, \textit{RegionTrans} can be seen as its realization  for spatio-temporal prediction tasks. For a city, \textit{region} is a natural and semantically meaningful dividing. Matching is built on the available short period of service data, or a long period of correlated auxiliary data (if applicable), to make the matched inter-city region pairs share similar spatio-temporal patterns. Besides \textit{RegionTrans}, we believe that the \textit{divide-match-transfer} principle can further guide the design of transfer learning algorithms for more tasks beyond spatio-temporal prediction.

\section{Experiment: Crowd Flow Prediction}

In the experiment, we use crowd flow prediction,  an important type of spatio-temporal prediction tasks in urban computing \cite{zhang2017deep,zhang2016dnn}, to verify the effectiveness of \textit{RegionTrans}. 

\subsection{Settings}
\textbf{Datasets.} Following previous studies on crowd flow \cite{hoang2016fccf,zhang2017deep,zhang2016dnn}, we use bike flow data for evaluation. Three bike flow datasets collected from \textit{Washington D.C.}, \textit{Chicago}, and \textit{New York City} are used. Each dataset covers a two-year period (2015-2016). In all the cities, the center area of $20km\times 20km$ are selected as the studied area. The area is split to $20 \times 20$ regions (i.e., each region is $1km \times 1 km$). 
For each evaluation scenario, we choose one city as the source city and another as the target. We assume that the source city has all its historical crowd flow data, but only a limited period of crowd flow data exists in the target city (e.g., one day). The last two-month data is chosen for testing. 

\textbf{Metric}. The evaluation metric is root mean square error (RMSE). Same as \cite{zhang2017deep}, the reported RMSE is the average RMSE of inflow and outflow.

\textbf{Network Implementation}. Our network structure implemented in the experiment has two layers of ConvLSTM with $5\times 5$ filters and 32 hidden states, to generate $\mathbf{X}_t^{rep} \in \mathbb R^{20\times 20 \times 32}$. With $\mathbf{X}_t^{rep}$ as the input, there is one layer of Conv2D$_{1\times 1}$ with 32 hidden states, followed by another layer of Conv2D$_{1\times1}$ linking to the output crowd flow prediction. For the external context factors, e.g., temperature, wind speed, weather, and day type, we use the same feature extraction method as \cite{zhang2017deep} and obtain an external feature vector  with a length of 28. We also need to set $w$ in Eq.~\ref{eq:obj} to balance the optimization trade-off between representation difference and prediction error. We set $w$ to 0.75 as the default value. \textit{ADAM} is used as the optimization algorithm \cite{kingma2014adam}.


\textbf{Methods.} 
For \textit{RegionTrans}, we implement two variants:
\begin{itemize}
	\item \textit{RegionTrans (S-Match)}: learning the inter-city region matching function $\mathcal M$ only by the short period of the target city \textit{Service} data, i.e., crowd flow.
	\item \textit{RegionTrans (A-Match)}: learning the inter-city region matching function $\mathcal M$ by the long period of the \textit{Auxiliary} data, i.e., Foursquare check-in data. We use one-year check-in data as the auxiliary data since it is a useful indication of crowd flow, as shown in previous studies~\cite{yang2016participatory}. Note that we have collected check-in data from D.C. and Chicago, so \textit{RegionTrans (A-Match)} is available for the knowledge transfer between D.C. $\rightleftharpoons$ Chicago.
\end{itemize}

We compare \textit{RegionTrans} with two types of baselines. The first type only uses the short crowd data history of target city for training its prediction model:

\begin{itemize}
	\item \textit{ARIMA}: a widely-used time series prediction method in statistics \cite{hyndman2014forecasting}.
	\item \textit{DeepST} \cite{zhang2016dnn}: a deep spatio-temporal neural network based on convolutional network. The complete DeepST model has three components: \textit{closeness}, \textit{period}, and \textit{trend}. But the \textit{period} and \textit{trend} components can only be activated if the training data last for more than one day and seven days, respectively. Therefore, if the target city does not have enough data, we have to deactivate the corresponding components.
	\item \textit{ST-ResNet} \cite{zhang2017deep}: a deep spatio-temporal neural network based on residual network \cite{he2016deep}. Same as DeepST, ST-ResNet has three components. We then adapt ST-ResNet in the same way as DeepST in our experiments.
\end{itemize}

The second type trains a deep model on the source city data, and \textit{fine-tune} it with the target city data:
\begin{itemize}
	\item \textit{DeepST (FT)}: fine-tuned DeepST.
	\item \textit{ST-ResNet (FT)}: fine-tuned ST-ResNet.
\end{itemize}

As mentioned in Sec.~\ref{sub:network_structure}, DeepST and ST-ResNet predict the city crowd flow as a whole, and thus we cannot fine tune their models between two cities of different sizes. Therefore, to make the comparison possible, our experiment selects the same area size for two cities. Note that \textit{RegionTrans} is able to transfer knowledge between two cities of different sizes, and thus is more flexible.

\renewcommand{\arraystretch}{1.2}

\subsection{Results}

\begin{table}[t]
	\scriptsize
	\centering
	\caption{Evaluation results. The target city holds 1 or 3-day crowd flow historical data. \textit{RegionTrans (A-Match)} is available for D.C. $\rightleftharpoons$ Chicago as we have collected check-in data for Chicago and D.C.}
	\begin{tabular}{lcccccccc}
		\toprule
		& \multicolumn{2}{c}{\textbf{D.C.$\to$Chicago}} & \multicolumn{2}{c}{\textbf{Chicago$\to$D.C.}} & \multicolumn{2}{c}{\textbf{D.C.$\to$NYC}} & \multicolumn{2}{c}{\textbf{NYC$\to$D.C.}} \\
		\cmidrule(lr){2-3} \cmidrule(lr){4-5} \cmidrule(lr){6-7} \cmidrule(lr){8-9}
		& 1-day & 3-day  & 1-day & 3-day & 1-day & 3-day & 1-day & 3-day   \\ \midrule
		\textbf{Target Data Only} &  &  &  & & &   \\
		\textit{ARIMA} & 0.740 & 0.694  & 0.707 & 0.661 & 0.360 & 0.341  & 0.707 & 0.661\\
		\textit{DeepST} & 0.771 & 0.711 & 1.075 & 0.767 & 0.350 & 0.359 & 1.075 & 0.767 \\
		\textit{ST-ResNet} & 0.914 & 0.703  & 0.869& 0.738 & 0.376 & 0.349 & 0.869& 0.738 \\ \midrule
		\textbf{Source \& Target Data}&  &  &  & & &\\
		\textit{DeepST (FT)} & 0.652 & 0.611 & 0.672 & 0.619 & 0.363 & 0.369 & 0.713 & 0.711\\
		\textit{ST-ResNet (FT)} & 0.667 & 0.615 & 0.695 & 0.623 & 0.385 & 0.349 & 0.696 & 0.691  \\
		\textit{RegionTrans (S-Match)} &  0.605 &  0.594&  0.631 &  0.602 & 0.328 & 0.305 & 0.665 & 0.593\\
		\textit{RegionTrans (A-Match)} &  0.587 &  0.576 &  0.600 &  0.581 & / & /  & / & /\\
		\bottomrule
	\end{tabular}
	\label{tbl:results_inflow}
	\vspace{-1.5em}
\end{table}

\textbf{Comparison with baselines.} Table~\ref{tbl:results_inflow} shows our results for D.C. $\rightleftharpoons$ Chicago and D.C. $\rightleftharpoons$ NYC. In all the scenarios, \textit{RegionTrans} can consistently outperform the best baseline, where the largest improvement is reducing the prediction error by up to 10.7\%.  In particular, when the recorded history of the target city is shorter, the improvement of \textit{RegionTrans} is usually more significant. This indicates that the introduced inter-city similar-region pairs are valuable for transfer learning especially when target data is extremely scarce. Between two variants of \textit{RegionTrans}, \textit{RegionTrans (A-Match)} is better, as shown in D.C. $\rightleftharpoons$ Chicago. This implies that if an appropriate type of auxiliary data exists, it is possible to build a better inter-city region matching than only using the short period of the service data. If the auxiliary data is  unavailable, using the limited period of service data for region matching can still lead to the competitive variant \textit{RegionTrans (S-Match)}, which beats all the baselines significantly.

Another important observation is that \textit{RegionTrans} is more robust when transferring knowledge between two dissimilar cities than baselines. Between the three cities in the experiment, D.C. and Chicago are similar in population, while NYC has a much larger population. This indicates that the knowledge transfer between D.C. $\rightleftharpoons$ Chicago may be easier, while D.C. $\rightleftharpoons$ NYC could be harder. Our results also verify this as DeepST and ST-ResNet get large improvement by fine-tuning in D.C. $\rightleftharpoons$ Chicago; but in D.C. $\to$ NYC, \textit{negative transfer} appears for the fine-tuned DeepST and ST-ResNet, leading to even worse performance than ARIMA, indicating that directly transferring the whole city knowledge from D.C. to NYC is ineffective. In comparison, \textit{RegionTrans} consistently achieves lower error than all the baselines, verifying that the knowledge from D.C. can still be effectively transferred to NYC. The primary reason that \textit{RegionTrans} can avoid negative transfer is that although D.C. and NYC are  dissimilar in general, we can still find inter-city region pairs with similar spatio-temporal patterns (e.g., central business district) to facilitate cross-city knowledge transfer.




\begin{figure}[t]
	\centering
	\includegraphics[width= .3\linewidth]{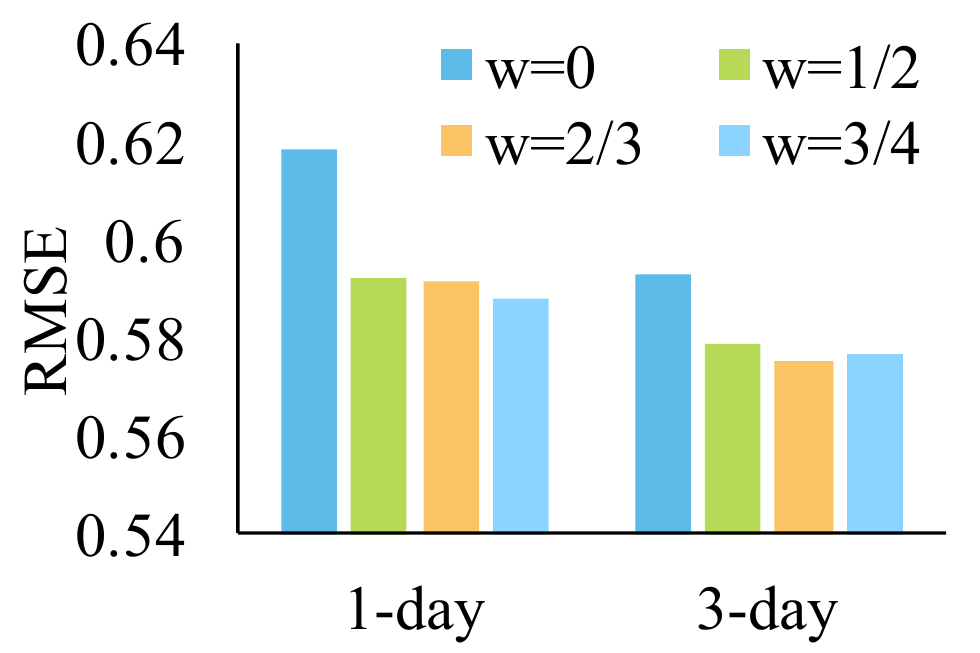}
	\caption{Tunning $w$ of RegionTrans (A-Match, D.C. $\to$ Chicago).}
	\label{fig:vary_w}
	\vspace{-1em}
\end{figure}

\textbf{Tuning $\bm{w}$.} We tune $w$ in Eq.~\ref{eq:obj} to see how it will affect the performance. The larger $w$ is,  the higher weight is put on minimizing the similar-region representation difference. Fig.~\ref{fig:vary_w} shows the results. If we set $w=0$, i.e., ignoring the inter-city similar-region representation in transfer learning, the performance is significantly worse than when $w>0$, by incurring up to 5\% higher error. This highlights the effectiveness of our proposed inter-city similar-region matching scheme in cross-city knowledge transfer. For other settings of $w>0$, the performance difference is minor. A larger $w$ performs slightly better when we have a very short period of target city crowd flow data, e.g., one day.



\textbf{Computation time.} The experiment platform is equipped with Intel Xeon CPU E5-2650L, 128 GB RAM, and Nvidia Tesla M60 GPU. We implement \textit{RegionTrans} with TensorFlow in CentOS. Training the source city model on two years of data needs about 20 minutes, and the transfer learning for the target city model costs about 50 and 100 minutes for 1 and 3-day data, respectively. This running time efficiency is acceptable in real-life deployments.

\section{Conclusion}

In this paper, to address the data scarcity issue in spatio-temporal  prediction tasks in urban computing, we propose a novel cross-city deep transfer learning framework, called \textit{RegionTrans}, with three stages. (1) We use a short period of service data, or a long period of auxiliary data if applicable, to obtain inter-city region similarities regarding the spatio-temporal dynamics. (2) We design a deep spatio-temporal model with a hidden layer dedicated to storing region-level latent representations. (3)~We propose a network parameter optimization algorithm to transfer knowledge from a source city to a target one by considering the latent representations of the inter-city similar-region pairs.

In the future, we plan to extend \textit{RegionTrans} in several directions. First, we will try to theoretically analyze the properties of \textit{RegionTrans} and the \textit{divide-match-transfer} principle behind it. Apparently, the transfer learning performance depends on how good the matching of the target regions and the source regions is; hence, mathematically modeling the relationship between the quality of matching and the final transfer performance will be our primary future work.
Second, we will consider a more general scenario where multiple data-rich source cities are available. Better performance may be achieved if we can combine transferable knowledge from multi-source cities.
Finally, we plan to extend \textit{RegionTrans} to spatio-temporal learning tasks besides prediction, e.g., facility deployment.

%

\small

\bibliographystyle{plain}
\bibliography{city_transfer_crowd}

\end{document}